\pdfoutput=1

\documentclass[11pt]{article}

\usepackage{ACL2023}
\usepackage{xcolor}
\usepackage{colortbl,xcolor}
\usepackage{booktabs}
\usepackage{multirow}
\usepackage{multicol}
\usepackage{times}
\usepackage{latexsym}
\usepackage{graphicx}
\usepackage{colortbl}
\usepackage{xcolor}
\usepackage[most]{tcolorbox}
\usepackage{xcolor}
\usepackage{pifont}
\newcommand{\cmark}{\ding{51}} 
\newcommand{\xmark}{\ding{55}} 
\usepackage[T1]{fontenc}

\usepackage[utf8]{inputenc}

\definecolor{darkred}{RGB}{255,150,150}     
\definecolor{lightred}{RGB}{255,200,200}    
\definecolor{darkyellow}{RGB}{255,230,150}  
\definecolor{lightyellow}{RGB}{255,255,200} 
\definecolor{lightgreen}{RGB}{220,255,220}  
\definecolor{darkgreen}{RGB}{180,255,180}   

\usepackage{microtype}

\usepackage{inconsolata}

\title{Modeling Contextual Passage Utility for Multihop Question Answering}

\author{
  Akriti Jain, Aparna Garimella \\
  Adobe Research, India \\
  \texttt{\{akritij, garimell\}@adobe.com}
}

\begin{document}
\maketitle
\begin{abstract}
Multihop Question Answering (QA) requires systems to identify and synthesize information from multiple text passages. While most prior retrieval methods assist in identifying relevant passages for QA, further assessing the utility of the passages can help in removing redundant ones, which may otherwise add to noise and inaccuracies in the generated answers. Existing utility prediction approaches model passage utility independently, overlooking a critical aspect of multi-hop reasoning, that the utility of a passage can be context-dependent, influenced by its relation to other passages—whether it provides complementary information, or forms a crucial link in conjunction with others. In this paper, we propose a light-weight approach to model contextual passage utility, accounting for inter-passage dependencies. We fine-tune a small transformer-based model to predict passage utility scores for multihop QA. We leverage the reasoning traces from an advanced reasoning model to capture the order in which passages are used to answer a question, to obtain synthetic training data. Through comprehensive experiments, we demonstrate that our utility-based scoring of retrieved passages leads to better reranking and downstream task performance compared to relevance-based reranking methods. 

\end{abstract}
\begin{figure}[t]
    \centering
    \includegraphics[width=0.45\textwidth]{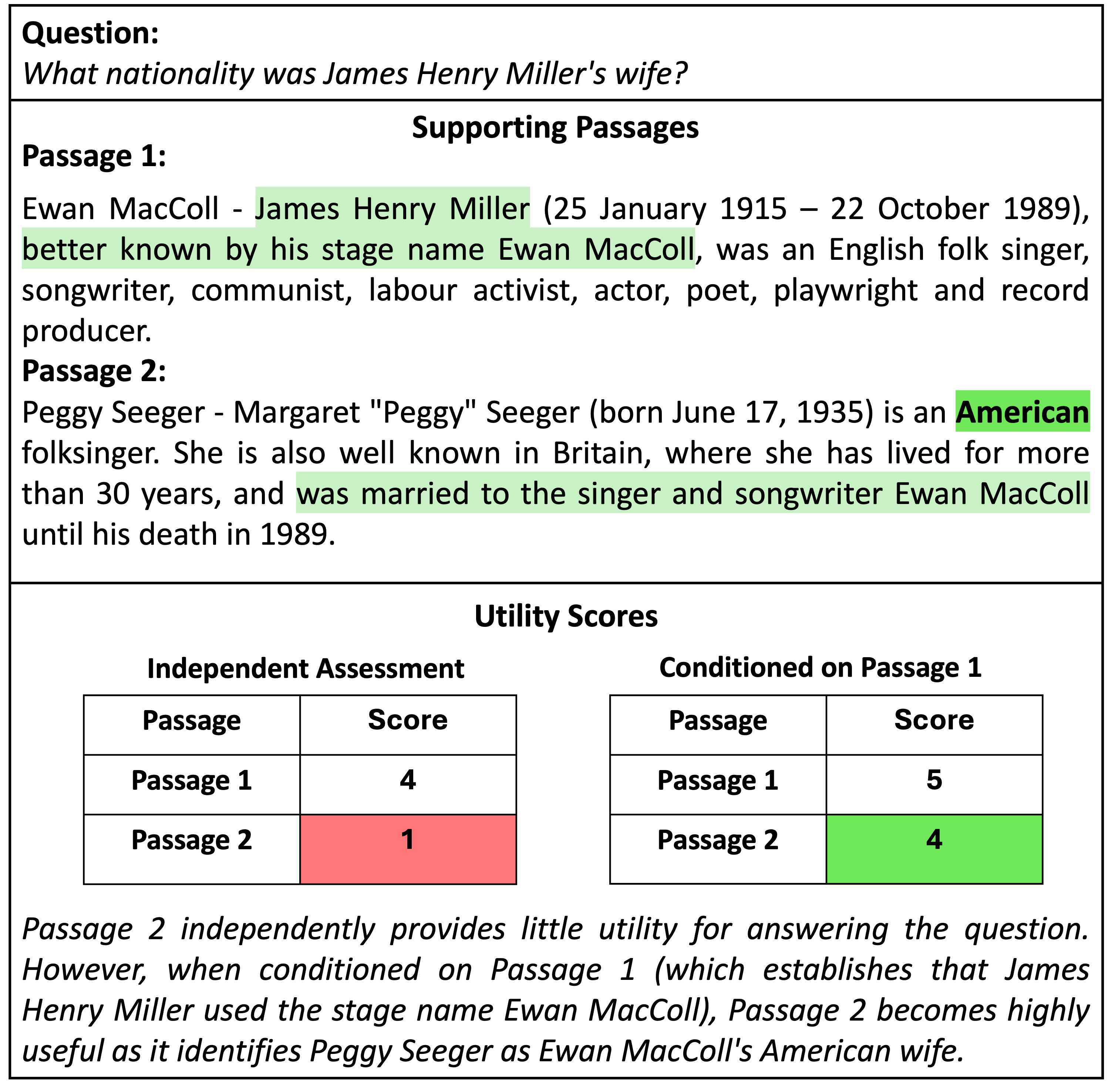}
    \caption{A multihop question from HotpotQA dataset \cite{yang2018hotpotqadatasetdiverseexplainable}: Passage 2 if considered independently does not seem useful to answer the question. However, conditioned on Passage 1, it becomes useful.}
    \label{fig:hero_example}
    \vspace{-0.15in}
\end{figure}
\section{Introduction}
Effective multihop question answering (QA) hinges on identifying not only relevant passages, but also those that are truly useful to answering the given question. While relevance merely signifies a topical connection between context and the query, utility reflects a passage’s actual contribution to constructing the answer \cite{xu2025trainingutilitybasedretrievershared}. 
Classic relevance labels often fail to predict QA success, and external relevance judgments tend to correlate poorly with QA performance \cite{salemievaluatingretrievalquality}. 
This insufficiency arises partly because retrieved passages may be individually relevant, yet only a subset of them are actually used within the specific inferential path required to reach the correct answer \cite{zhang2024largelanguagemodelsgood}. 
Consequently, utility judgments, rather than relevance alone, offer more valuable guidance for identifying ground-truth evidence and enhancing answer generation.


The work on passage utility modeling has been nascent.
\citet{perezbeltrachini2025uncertaintyquantificationretrievalaugmented} proposed to incorporate factors such as QA model's accuracy and entailment scores between the passages and the answer, to predict the utility of a given passage. 
Despite the growing recognition, existing methods fall short in multihop QA. 
A main limitation is the tendency to assess passages in isolation, assuming their contribution is independent of other contextual information. 
This is particularly problematic in multihop scenarios where a passage's utility is frequently conditional—it may only become useful after another passage has established necessary context, such as a supporting entity in one passage activating the usefulness of subsequent information (Figure \ref{fig:hero_example}). 
Thus, compositionality and order are paramount: a passage's utility is to be determined by its position within a reasoning chain and its dependencies on preceding content. 

Our work aims to address this gap by proposing a novel approach to model passage utility by accounting for inter-passage dependencies.
We synthetically generate utility ratings for passages, using the reasoning traces from an advanced reasoning model (o1) to capture the order in which the passages are used in addressing a question, which are then used for point-wise passage utility ratings on a scale of 1-5 annotated by GPT-4o. 
A RoBERTa-based model \cite{liu2019roberta} is trained on these trace-level annotations to explicitly learn these contextual dependencies to predict utility scores. 

Our contributions are threefold: 
{\bf (1)} We address utility-aware passage ranking in multihop QA, which remains an underexplored topic in retrieval-augmented QA efforts.
{\bf (2)} We propose a light-weight scoring model that learns to predict utility, sensitive to inter-passage dependencies. We obtain path-dependent utility in multi-hop QA using ordered reasoning traces and LLM-generated ordinal scores.
{\bf (3)} Through experiments on various datasets, we illustrate that our approach significantly improves the identification of useful passage sets and enhances downstream QA performance compared to baselines that do not account for utility, or do so in a context independent manner.

\section{Utility-Aware Contextual Passage Ranking}
We formulate passage utility scoring in multihop question answering as a regression task. 
We maintain that the true utility of a passage $p_i$ is inherently context-dependent, influenced by previously encountered passages $P_{<i}$ and the question $q$. 
Our objective is to predict a utility score $U(p_i | P_{<i}, q)$ on a scale of 1 to 5. 
A high score signifies a greater contribution of $p_i$ towards answering $q$, given the contextual information derived from $P_{<i}$. Here, $i$ indexes the position of $p_i$ in an ordered set of retrieved passages.

Our passage utility predictor uses RoBERTa-large \cite{liu2019roberta} as a regression model. It takes as input the question $q$ and the current passage $p_i$ being evaluated, and predicts a utility score indicating how useful $p_i$ is for answering $q$.
Our pipeline comprises of two stages: (1) synthetic data generation with utility scores for a given passage, conditioned on the other context, for a given question; and (2) utility predictor training using the RoBERTa-large model.

\subsection{Training Data Generation}
Our training data consists of (question, passage, utility score) triplets, where the utility scores are derived through a two-staged process designed to capture contextual dependencies. With the growing use of LLMs for generating high-quality synthetic training data \cite{wagner2025powerllmgeneratedsyntheticdata, li2023syntheticdatagenerationlarge,ba2024gapsmodelcalibrationgeneralization}, we adopt a similar strategy, allowing the reasoning capabilities of powerful LLMs to be distilled into smaller, more efficient models for downstream tasks.\\
\noindent
\textbf{Reasoning Trace Generation.}
To capture the compositional nature of information in multi-hop reasoning, we first generate explicit reasoning traces using a pre-trained reasoning model \cite{openai2024openaio1card}. Each passage within the context for a given question is tagged with a unique identifier (e.g., [Passage A], [Passage B]). The reasoning model is then prompted to produce a detailed reasoning trace that explicitly cites which passage(s) support each inferential step required to answer the question. 
These reasoning traces clarify how information is progressively integrated across multiple passages and illustrate the specific utility each passage contributes toward constructing the final answer in a multi-hop setting. \\
\noindent
\textbf{Utility Score Annotation.}
We employ another LLM, GPT-4o, to assign utility scores (ranging from 1 to 5) to each passage, conditioned on its function within the complete reasoning trace generated in the previous step.
For each passage $p_i$, the scoring LLM is provided with the question $q$, the target passage $p_i$, and the full reasoning trace $T$ associated with $q$. 
The LLM is prompted to evaluate $p_i$'s utility by analyzing its usage in $T$. Specifically, it assesses: (1) whether $p_i$ was explicitly cited as evidence in $T$; (2) the specific role $p_i$ played according to $T$ (e.g., providing initial facts, bridging information, or supplying final evidence); and (3) the criticality of this contribution to answering $q$ as per $T$. 
The utility is then quantified on a 5-point scale, where 1 indicates a passage not being used to address $q$, and 5 represents an essential passage providing critical evidence without which the answer could not be derived according to the trace. This approach is to ensure that the utility scores $U_i$ reflect not just the intrinsic relevance of a passage to the question, but its actual contribution within the specific reasoning context. These context-aware utility scores serve as supervision signals for training our regression model and their reliability is supported by a high upper-bound performance when used directly for passage ranking.
We manually verify 200 examples across our different datasets;
we note alignment between LLM-generated and human-rated passage utilities in over $95\%$ cases, confirming their reliability as supervision signals (further details on both in Appendix~\ref{sec:appendix-verification}).

\subsection{Training Procedure}
We fine-tune the RoBERTa-large model by adding a sequence classification head configured for regression, which outputs a single scalar value representing the predicted utility score.  While the RoBERTa model itself only receives the local $(q, p_i)$ pair as direct input, it is trained to predict the context-dependent utility scores $U_i$ (derived from the full reasoning trace), by minimizing the mean squared error (MSE) between its predicted utility scores $f_{RoBERTa}(p_i, q)$ and the LLM-annotated, context-dependent scores $U_i$.
\begin{equation}
L = \frac{1}{N} \sum_{j=1}^{N} (f_{RoBERTa}(p_{i,j}, q_j) - U_{i,j})^2
\end{equation}

\noindent where $N$ is the total number of training examples.
Here, $j$ indexes individual training examples, each consisting of a question $q_j$, a passage $p_{i,j}$, and a utility score $U_{i,j}$.

\section{Experimental Setup}
We fine-tune the RoBERTa model for 3 epochs using the Adam optimizer \cite{kingma2017adammethodstochasticoptimization} with a weight decay of 0.01, along with a linear learning rate scheduler and a warmup ratio of 0.1. 
Training is performed with a per-device batch size of 8 and gradient accumulation steps set to 2. 
Mixed-precision (FP16) training is enabled. 
All experiments are conducted on a single NVIDIA A100 GPU (40GB).

\subsection{Datasets}
We evaluate our contextual passage utility scorer on three multihop QA benchmarks. 

\noindent\textbf{HotpotQA} \cite{yang2018hotpotqadatasetdiverseexplainable} features questions requiring 2-hop reasoning. Each question is typically accompanied by 2 gold supporting passages and 8 distractor passages. \\
\noindent
\textbf{MusiQue} \cite{trivedi2022musiquemultihopquestionssinglehop} is designed for compositional reasoning. The questions involve 2 to 4 reasoning hops. Each question includes 20 passages in total.\\
\noindent
\textbf{2WikiMultiHopQA} \cite{ho2020constructingmultihopqadataset} contains 2 or 4-hop questions constructed from Wikipedia, requiring the model to combine information from two different Wikipedia entities or facts to arrive at the answer. The total number of passages are 10.

\noindent
The datasets we evaluate on already provide a fixed set of supporting and distractor passages for each question. This setup can be viewed as the output of a first-stage retriever, and hence, the number of candidate passages that our model scores is determined by each dataset’s official construction.

\noindent
We randomly select 5K question-passage pairs \((q; p_1, p_2, \ldots)\) for training, each annotated with a utility score from 1 to 5, based on the above setup.

\begin{table*}[htbp]
\centering
\scalebox{0.75}{
\setlength{\tabcolsep}{4pt}
\begin{tabular}{llcccccccccc}
\toprule
\textbf{Dataset} & \textbf{Method} & \textbf{Multi-hop} & \textbf{P@2} & \textbf{R@2} & \textbf{F1@2} & \textbf{R@5} & \textbf{NDCG@1} & \textbf{NDCG@5} & \textbf{EM} \\
\midrule

\multirow{7}{*}{\textbf{HotpotQA}} 
& BM25*        & \xmark & 52.32 & 52.32 & 52.32 & 73.22 & 70.27 & 56.39 & 54.41 \\
& Contriever* & \xmark & 47.31 & 47.31 & 47.31 & 74.38 & 59.03 & 64.89 & 49.80 \\
& MonoT5      & \xmark & 40.86 & 40.86 & 40.86 & 72.34 & 40.99 & 40.89 & 48.56 \\
& MDR*         & \cmark & 62.04 & 62.04 & 62.04 & 85.62 & 80.05 & \cellcolor{lightred}79.24 & 73.57 \\
& PromptRank  & \cmark & 50.79 & 50.79 & 50.79 & 71.01 & 70.28 & 66.31 & 61.05 \\
& LLM         & \cmark & \cellcolor{lightred}70.20 & \cellcolor{lightred}70.20 & \cellcolor{lightred}70.20 & \cellcolor{lightred}86.10 & \cellcolor{lightred}85.22 & 73.49 & \cellcolor{lightred}84.86 \\
\hline
& \textbf{Ours}        & \cmark & \cellcolor{lightgreen}\textbf{88.09} & \cellcolor{lightgreen}\textbf{88.09} & \cellcolor{lightgreen}\textbf{88.09} & \cellcolor{lightgreen}\textbf{98.33} & \cellcolor{lightgreen}\textbf{94.61} & \cellcolor{lightgreen}\textbf{89.57} & \cellcolor{lightgreen}\textbf{87.12} \\
\midrule

\multirow{7}{*}{\textbf{MuSiQue}} 
& BM25*        & \xmark & 34.05 & 25.70 & 29.29 & 44.06 & 44.19 & 36.34 & 39.47 \\
& Contriever* & \xmark & 31.53 & 23.80 & 27.12 & 42.66 & 38.39 & 38.83 & 32.89 \\
& MonoT5      & \xmark & \cellcolor{lightred}51.70 & 39.02 & 44.47 & 63.11 & 60.74 & 53.74 & 55.11 \\
& MDR*         & \cmark & 51.11 & \cellcolor{lightred}41.19 & \cellcolor{lightred}45.62 & \cellcolor{lightred}66.65 & 63.01 & \cellcolor{lightred}61.75 & 61.30 \\
& PromptRank  & \cmark & 42.20 & 32.90 & 36.99 & 48.84 & 57.84 & 48.23 & 32.77 \\
& LLM         & \cmark & 45.51 & 37.04 & 40.17 & 50.06 & \cellcolor{lightred}64.36 & 50.06 & \cellcolor{lightgreen}\textbf{68.97} \\
\hline
& \textbf{Ours}        & \cmark & \cellcolor{lightgreen}\textbf{62.37} & \cellcolor{lightgreen}\textbf{47.08} & \cellcolor{lightgreen}\textbf{53.66} & \cellcolor{lightgreen}\textbf{73.23} & \cellcolor{lightgreen}\textbf{78.94} & \cellcolor{lightgreen}\textbf{66.12} & \cellcolor{lightred}68.51 \\
\midrule

\multirow{7}{*}{\textbf{2WikiMultiHopQA}} 
& BM25*        & \xmark & 49.75 & 40.82 & 44.85 & 67.68 & 61.84 & 52.49 & 32.26 \\
& Contriever* & \xmark & 26.86 & 22.04 & 24.21 & 41.48 & 33.97 & 36.21 & 18.38 \\
& MonoT5      & \xmark & 61.70 & 50.63 & 55.62 & \cellcolor{lightred}80.79 & 65.60 & 62.59 & 50.85 \\
& MDR*         & \cmark & \cellcolor{lightred}70.45 & \cellcolor{lightred}61.29 & \cellcolor{lightred}65.55 & 66.65 & 63.01 & 61.75 & \cellcolor{lightred}70.26 \\
& PromptRank  & \cmark & 48.35 & 43.05 & 45.54 & 62.20 & 65.31 & 59.32 & 40.55 \\
& LLM         & \cmark & 62.50 & 54.51 & 57.18 & 75.10 & \cellcolor{lightred}75.93 & \cellcolor{lightred}65.57 & 61.75 \\
\hline
& \textbf{Ours}        & \cmark & \cellcolor{lightgreen}\textbf{94.51} & \cellcolor{lightgreen}\textbf{83.66} & \cellcolor{lightgreen}\textbf{88.75} & \cellcolor{lightgreen}\textbf{99.02} & \cellcolor{lightgreen}\textbf{97.70} & \cellcolor{lightgreen}\textbf{95.23} & \cellcolor{lightgreen}\textbf{79.44} \\
\bottomrule
\end{tabular}
}
\caption{Evaluation of contextual passage utility scoring on multi-hop QA datasets. EM = Exact Match; P@2 = Precision@2; R@2 = Recall@2; R@5 = Recall@5. \textcolor{lightgreen}{Green} highlights best performance; \textcolor{lightred}{Red} highlights second-best. * indicates retriever-based methods.}

\label{tab:contextual-utility-eval}
\end{table*}

\subsection{Evaluation Metrics}
\label{sec:evaluation_metrics} 

Our evaluation targets three key aspects: identifying the correct set of ground-truth passages (\(P_G\)), ranking them effectively, and their utility in answering the question using the model-selected passages (\(P_M\)). For rank-based metrics, we consider the top 5 selected passages ($k=5$). To assess how well \(P_M\) covers \(P_G\), we use Precision@k (P@k), the fraction of selected passages that are correct; Recall@k (R@k), the fraction of ground-truth passages retrieved; and F1-Score@k (F1@k), the harmonic mean of P@k and R@k.
Next, to evaluate the ranking quality of ground-truth passages within the top 5, we employ Normalized Discounted Cumulative Gain (NDCG@1, NDCG@5) \cite{10.1145/3130348.3130374}, which measures overall ranking quality at the top 1 and top 5 positions. 
For downstream task performance, we assess if the top-k passages selected by our model are sufficient to answer the question. We report Exact Match (EM), which measures the percentage of predicted answers (generated using \(P_M\)) that exactly match a gold answer.

\subsection{Baselines}

We compare our model against several retrieval and reranking approaches. 
All rerankers operate on the same set of candidate passages per question. 1) \textbf{BM25:} A classical sparse retriever \cite{10.1561/1500000019} that uses term-based scoring to rank candidate passages. 2) \textbf{Contriever:} An unsupervised dense retriever \cite{izacard2021contriever}. We use its raw similarity scores for ranking, representing a strong dense retrieval baseline. 3) \textbf{MonoT5:} A T5-based neural reranker \cite{nogueira2020documentrankingpretrainedsequencetosequence}. We use a pre-trained model (castorini/monot5-base-msmarco) for pointwise passage relevance scoring. 4) \textbf{LLM-based Reranker (Zero-Shot):} GPT-4o prompted in a zero-shot manner to score the relevance of each candidate passage to the question. 3) Cross-Attention Reranking is a post-retrieval step integrated with the \textbf{Multi-Hop Dense Retrieval (MDR) }\cite{xiong2021answeringcomplexopendomainquestions} system. It takes the top-k passage sequences retrieved by MDR, prepends the original question to each, and uses a pre-trained Transformer encoder (like ELECTRA-large) to predict relevance scores. 5) \textbf{PROMPTRANK} \cite{kong2023promptrankunsupervisedkeyphraseextraction} is an LLM-based reranker designed for few-shot multi-hop QA. It scores candidate document paths (sequences of documents) by measuring the large language model's conditional likelihood of generating the question given a prompt constructed from that path.
\section{Results \& Discussion}
Table 1 shows the results across the three datasets. Our approach consistently outperforms all the baselines in both identifying the full set of useful passages (coverage - high R@k) and correctly ordering them (ranking - high NDCG scores). We observe significant improvements over even strong multi-hop–aware rerankers like MDR and PromptRank.
On HotpotQA, for instance, our model achieves R@5 of $98.33$, substantially higher than the strongest baseline (LLM scorer at $86.10$), and NDCG@5 of $89.57$ compared to the MDR Reranker $79.24$. 
Similar gains are observed on MuSiQue and 2WikiMultiHopQA datasets as well.

\noindent\textbf{Auxiliary Experiment.}
To further validate our approach, we conduct an auxiliary experiment with two key findings (detailed in Appendix~\ref{sec:slm}). In this setup, we fine-tune decoder-only models (LLaMA 3.2 1B and LLaMA 3.1 8B) to predict passage utility scores in two settings: (i) listwise scoring, where all candidate passages for a question are provided jointly, and (ii) pointwise scoring, where each passage is scored independently. Providing the full joint context in the listwise setup yields only marginal improvements over pointwise scoring. In contrast, our RoBERTa-based model, despite scoring one passage at a time, consistently outperforms these larger LLaMA models even when they have access to all candidate passages simultaneously. This indicates that the essential cross-passage dependencies are effectively distilled into the learned utility scores themselves. Encoder-based models, being lightweight and inherently suited for regression, are thus better positioned to leverage this distilled signal, explaining their strong performance in state-of-the-art reranking systems. Further research on leveraging this utility regression capabilities demonstrated by encoders while simultaneously benefiting from the comprehensive contextual understanding offered by large-context decoders would be beneficial in more complex open-domain query-based generation tasks.

\newcommand{\gain}[1]{\textcolor{green!60!black}{\scriptsize\,(+{#1}\%)}}
\newcommand{\drop}[1]{\textcolor{red!75!black}{\scriptsize\,($-${#1}\%)}}

\section*{Limitations}
Current multi-hop datasets are predominantly fact-based. Consequently, a model trained on such data may not generalize effectively to tasks requiring high-level inferential reasoning. For example: answering questions like, "Why was the publication of The Catcher in the Rye considered provocative when it was released?". In these scenarios, the \textit{utility} of a passage might be tied to more abstract semantic properties.

\section*{Ethics Statement}
There are no ethical concerns to the best of our knowledge.

\bibliography{anthology,custom}

@misc{xu2025trainingutilitybasedretrievershared,
      title={Training a Utility-based Retriever Through Shared Context Attribution for Retrieval-Augmented Language Models}, 
      author={Yilong Xu and Jinhua Gao and Xiaoming Yu and Yuanhai Xue and Baolong Bi and Huawei Shen and Xueqi Cheng},
      year={2025},
      eprint={2504.00573},
      archivePrefix={arXiv},
      primaryClass={cs.CL},
      url={https://arxiv.org/abs/2504.00573}, 
}

@misc{perezbeltrachini2025uncertaintyquantificationretrievalaugmented,
      title={Uncertainty Quantification in Retrieval Augmented Question Answering}, 
      author={Laura Perez-Beltrachini and Mirella Lapata},
      year={2025},
      eprint={2502.18108},
      archivePrefix={arXiv},
      primaryClass={cs.CL},
      url={https://arxiv.org/abs/2502.18108}, 
}

@misc{zhang2024largelanguagemodelsgood,
      title={Are Large Language Models Good at Utility Judgments?}, 
      author={Hengran Zhang and Ruqing Zhang and Jiafeng Guo and Maarten de Rijke and Yixing Fan and Xueqi Cheng},
      year={2024},
      eprint={2403.19216},
      archivePrefix={arXiv},
      primaryClass={cs.IR},
      url={https://arxiv.org/abs/2403.19216}, 
}

@misc{yang2018hotpotqadatasetdiverseexplainable,
      title={HotpotQA: A Dataset for Diverse, Explainable Multi-hop Question Answering}, 
      author={Zhilin Yang and Peng Qi and Saizheng Zhang and Yoshua Bengio and William W. Cohen and Ruslan Salakhutdinov and Christopher D. Manning},
      year={2018},
      eprint={1809.09600},
      archivePrefix={arXiv},
      primaryClass={cs.CL},
      url={https://arxiv.org/abs/1809.09600}, 
}

@misc{ho2020constructingmultihopqadataset,
      title={Constructing A Multi-hop QA Dataset for Comprehensive Evaluation of Reasoning Steps}, 
      author={Xanh Ho and Anh-Khoa Duong Nguyen and Saku Sugawara and Akiko Aizawa},
      year={2020},
      eprint={2011.01060},
      archivePrefix={arXiv},
      primaryClass={cs.CL},
      url={https://arxiv.org/abs/2011.01060}, 
}

@misc{trivedi2022musiquemultihopquestionssinglehop,
      title={MuSiQue: Multihop Questions via Single-hop Question Composition}, 
      author={Harsh Trivedi and Niranjan Balasubramanian and Tushar Khot and Ashish Sabharwal},
      year={2022},
      eprint={2108.00573},
      archivePrefix={arXiv},
      primaryClass={cs.CL},
      url={https://arxiv.org/abs/2108.00573}, 
}

@inproceedings{liu2019roberta,
  title={RoBERTa: A Robustly Optimized BERT Pretraining Approach},
  author={Liu, Yinhan and Ott, Myle and Goyal, Naman and Du, Jingfei and Joshi, Mandar and Chen, Danqi and Levy, Omer and Lewis, Mike and Zettlemoyer, Luke and Stoyanov, Veselin},
  booktitle={Proceedings of the 7th International Conference on Learning Representations},
  year={2019}
}

@inproceedings{salemievaluatingretrievalquality,
author = {Salemi, Alireza and Zamani, Hamed},
title = {Evaluating Retrieval Quality in Retrieval-Augmented Generation},
year = {2024},
isbn = {9798400704314},
publisher = {Association for Computing Machinery},
address = {New York, NY, USA},
url = {https://doi.org/10.1145/3626772.3657957},
doi = {10.1145/3626772.3657957},
booktitle = {Proceedings of the 47th International ACM SIGIR Conference on Research and Development in Information Retrieval},
pages = {2395–2400},
numpages = {6},
location = {Washington DC, USA},
series = {SIGIR '24}
}

@misc{izacard2021contriever,
      title={Unsupervised Dense Information Retrieval with Contrastive Learning}, 
      author={Gautier Izacard and Mathilde Caron and Lucas Hosseini and Sebastian Riedel and Piotr Bojanowski and Armand Joulin and Edouard Grave},
      year={2021},
      url = {https://arxiv.org/abs/2112.09118},
      doi = {10.48550/ARXIV.2112.09118},
}

@misc{nogueira2020documentrankingpretrainedsequencetosequence,
      title={Document Ranking with a Pretrained Sequence-to-Sequence Model}, 
      author={Rodrigo Nogueira and Zhiying Jiang and Jimmy Lin},
      year={2020},
      eprint={2003.06713},
      archivePrefix={arXiv},
      primaryClass={cs.IR},
      url={https://arxiv.org/abs/2003.06713}, 
}

@misc{openai2024openaio1card,
      title={OpenAI o1 System Card}, 
      author={OpenAI and : and Aaron Jaech and Adam Kalai and Adam Lerer and Adam Richardson and Ahmed El-Kishky and Aiden Low and Alec Helyar and Aleksander Madry and Alex Beutel and Alex Carney and Alex Iftimie and Alex Karpenko and Alex Tachard Passos and Alexander Neitz and Alexander Prokofiev and Alexander Wei and Allison Tam and Ally Bennett and Ananya Kumar and Andre Saraiva and Andrea Vallone and Andrew Duberstein and Andrew Kondrich and Andrey Mishchenko and Andy Applebaum and Angela Jiang and Ashvin Nair and Barret Zoph and Behrooz Ghorbani and Ben Rossen and Benjamin Sokolowsky and Boaz Barak and Bob McGrew and Borys Minaiev and Botao Hao and Bowen Baker and Brandon Houghton and Brandon McKinzie and Brydon Eastman and Camillo Lugaresi and Cary Bassin and Cary Hudson and Chak Ming Li and Charles de Bourcy and Chelsea Voss and Chen Shen and Chong Zhang and Chris Koch and Chris Orsinger and Christopher Hesse and Claudia Fischer and Clive Chan and Dan Roberts and Daniel Kappler and Daniel Levy and Daniel Selsam and David Dohan and David Farhi and David Mely and David Robinson and Dimitris Tsipras and Doug Li and Dragos Oprica and Eben Freeman and Eddie Zhang and Edmund Wong and Elizabeth Proehl and Enoch Cheung and Eric Mitchell and Eric Wallace and Erik Ritter and Evan Mays and Fan Wang and Felipe Petroski Such and Filippo Raso and Florencia Leoni and Foivos Tsimpourlas and Francis Song and Fred von Lohmann and Freddie Sulit and Geoff Salmon and Giambattista Parascandolo and Gildas Chabot and Grace Zhao and Greg Brockman and Guillaume Leclerc and Hadi Salman and Haiming Bao and Hao Sheng and Hart Andrin and Hessam Bagherinezhad and Hongyu Ren and Hunter Lightman and Hyung Won Chung and Ian Kivlichan and Ian O'Connell and Ian Osband and Ignasi Clavera Gilaberte and Ilge Akkaya and Ilya Kostrikov and Ilya Sutskever and Irina Kofman and Jakub Pachocki and James Lennon and Jason Wei and Jean Harb and Jerry Twore and Jiacheng Feng and Jiahui Yu and Jiayi Weng and Jie Tang and Jieqi Yu and Joaquin Quiñonero Candela and Joe Palermo and Joel Parish and Johannes Heidecke and John Hallman and John Rizzo and Jonathan Gordon and Jonathan Uesato and Jonathan Ward and Joost Huizinga and Julie Wang and Kai Chen and Kai Xiao and Karan Singhal and Karina Nguyen and Karl Cobbe and Katy Shi and Kayla Wood and Kendra Rimbach and Keren Gu-Lemberg and Kevin Liu and Kevin Lu and Kevin Stone and Kevin Yu and Lama Ahmad and Lauren Yang and Leo Liu and Leon Maksin and Leyton Ho and Liam Fedus and Lilian Weng and Linden Li and Lindsay McCallum and Lindsey Held and Lorenz Kuhn and Lukas Kondraciuk and Lukasz Kaiser and Luke Metz and Madelaine Boyd and Maja Trebacz and Manas Joglekar and Mark Chen and Marko Tintor and Mason Meyer and Matt Jones and Matt Kaufer and Max Schwarzer and Meghan Shah and Mehmet Yatbaz and Melody Y. Guan and Mengyuan Xu and Mengyuan Yan and Mia Glaese and Mianna Chen and Michael Lampe and Michael Malek and Michele Wang and Michelle Fradin and Mike McClay and Mikhail Pavlov and Miles Wang and Mingxuan Wang and Mira Murati and Mo Bavarian and Mostafa Rohaninejad and Nat McAleese and Neil Chowdhury and Neil Chowdhury and Nick Ryder and Nikolas Tezak and Noam Brown and Ofir Nachum and Oleg Boiko and Oleg Murk and Olivia Watkins and Patrick Chao and Paul Ashbourne and Pavel Izmailov and Peter Zhokhov and Rachel Dias and Rahul Arora and Randall Lin and Rapha Gontijo Lopes and Raz Gaon and Reah Miyara and Reimar Leike and Renny Hwang and Rhythm Garg and Robin Brown and Roshan James and Rui Shu and Ryan Cheu and Ryan Greene and Saachi Jain and Sam Altman and Sam Toizer and Sam Toyer and Samuel Miserendino and Sandhini Agarwal and Santiago Hernandez and Sasha Baker and Scott McKinney and Scottie Yan and Shengjia Zhao and Shengli Hu and Shibani Santurkar and Shraman Ray Chaudhuri and Shuyuan Zhang and Siyuan Fu and Spencer Papay and Steph Lin and Suchir Balaji and Suvansh Sanjeev and Szymon Sidor and Tal Broda and Aidan Clark and Tao Wang and Taylor Gordon and Ted Sanders and Tejal Patwardhan and Thibault Sottiaux and Thomas Degry and Thomas Dimson and Tianhao Zheng and Timur Garipov and Tom Stasi and Trapit Bansal and Trevor Creech and Troy Peterson and Tyna Eloundou and Valerie Qi and Vineet Kosaraju and Vinnie Monaco and Vitchyr Pong and Vlad Fomenko and Weiyi Zheng and Wenda Zhou and Wes McCabe and Wojciech Zaremba and Yann Dubois and Yinghai Lu and Yining Chen and Young Cha and Yu Bai and Yuchen He and Yuchen Zhang and Yunyun Wang and Zheng Shao and Zhuohan Li},
      year={2024},
      eprint={2412.16720},
      archivePrefix={arXiv},
      primaryClass={cs.AI},
      url={https://arxiv.org/abs/2412.16720}, 
}

@article{10.1145/3130348.3130374,
author = {J\"{a}rvelin, Kalervo and Kek\"{a}l\"{a}inen, Jaana},
title = {IR evaluation methods for retrieving highly relevant documents},
year = {2017},
issue_date = {July 2017},
publisher = {Association for Computing Machinery},
address = {New York, NY, USA},
volume = {51},
number = {2},
issn = {0163-5840},
url = {https://doi.org/10.1145/3130348.3130374},
doi = {10.1145/3130348.3130374},
journal = {SIGIR Forum},
month = aug,
pages = {243–250},
numpages = {8}
}

@article{10.1561/1500000019,
author = {Robertson, Stephen and Zaragoza, Hugo},
title = {The Probabilistic Relevance Framework: BM25 and Beyond},
year = {2009},
issue_date = {April 2009},
publisher = {Now Publishers Inc.},
address = {Hanover, MA, USA},
volume = {3},
number = {4},
issn = {1554-0669},
url = {https://doi.org/10.1561/1500000019},
doi = {10.1561/1500000019},
journal = {Found. Trends Inf. Retr.},
month = apr,
pages = {333–389},
numpages = {57}
}

@article{meta2024llama3,
  title={The Llama 3 Herd of Models},
  author={Meta},
  journal={arXiv preprint arXiv:2407.21783},
  year={2024},
  url={https://arxiv.org/abs/2407.21783}
}

@misc{kong2023promptrankunsupervisedkeyphraseextraction,
      title={PromptRank: Unsupervised Keyphrase Extraction Using Prompt}, 
      author={Aobo Kong and Shiwan Zhao and Hao Chen and Qicheng Li and Yong Qin and Ruiqi Sun and Xiaoyan Bai},
      year={2023},
      eprint={2305.04490},
      archivePrefix={arXiv},
      primaryClass={cs.IR},
      url={https://arxiv.org/abs/2305.04490}, 
}

@misc{xiong2021answeringcomplexopendomainquestions,
      title={Answering Complex Open-Domain Questions with Multi-Hop Dense Retrieval}, 
      author={Wenhan Xiong and Xiang Lorraine Li and Srini Iyer and Jingfei Du and Patrick Lewis and William Yang Wang and Yashar Mehdad and Wen-tau Yih and Sebastian Riedel and Douwe Kiela and Barlas Oğuz},
      year={2021},
      eprint={2009.12756},
      archivePrefix={arXiv},
      primaryClass={cs.CL},
      url={https://arxiv.org/abs/2009.12756}, 
}

@misc{kingma2017adammethodstochasticoptimization,
      title={Adam: A Method for Stochastic Optimization}, 
      author={Diederik P. Kingma and Jimmy Ba},
      year={2017},
      eprint={1412.6980},
      archivePrefix={arXiv},
      primaryClass={cs.LG},
      url={https://arxiv.org/abs/1412.6980}, 
}

@misc{li2023syntheticdatagenerationlarge,
      title={Synthetic Data Generation with Large Language Models for Text Classification: Potential and Limitations}, 
      author={Zhuoyan Li and Hangxiao Zhu and Zhuoran Lu and Ming Yin},
      year={2023},
      eprint={2310.07849},
      archivePrefix={arXiv},
      primaryClass={cs.CL},
      url={https://arxiv.org/abs/2310.07849}, 
}

@misc{wagner2025powerllmgeneratedsyntheticdata,
      title={The Power of LLM-Generated Synthetic Data for Stance Detection in Online Political Discussions}, 
      author={Stefan Sylvius Wagner and Maike Behrendt and Marc Ziegele and Stefan Harmeling},
      year={2025},
      eprint={2406.12480},
      archivePrefix={arXiv},
      primaryClass={cs.CL},
      url={https://arxiv.org/abs/2406.12480}, 
}

@misc{ba2024gapsmodelcalibrationgeneralization,
      title={Fill In The Gaps: Model Calibration and Generalization with Synthetic Data}, 
      author={Yang Ba and Michelle V. Mancenido and Rong Pan},
      year={2024},
      eprint={2410.10864},
      archivePrefix={arXiv},
      primaryClass={cs.CL},
      url={https://arxiv.org/abs/2410.10864}, 
}

@misc{baker2024lostmiddleinbetweenenhancing,
      title={Lost in the Middle, and In-Between: Enhancing Language Models' Ability to Reason Over Long Contexts in Multi-Hop QA}, 
      author={George Arthur Baker and Ankush Raut and Sagi Shaier and Lawrence E Hunter and Katharina von der Wense},
      year={2024},
      eprint={2412.10079},
      archivePrefix={arXiv},
      primaryClass={cs.CL},
      url={https://arxiv.org/abs/2412.10079}, 
}

@misc{liu2023lostmiddlelanguagemodels,
      title={Lost in the Middle: How Language Models Use Long Contexts}, 
      author={Nelson F. Liu and Kevin Lin and John Hewitt and Ashwin Paranjape and Michele Bevilacqua and Fabio Petroni and Percy Liang},
      year={2023},
      eprint={2307.03172},
      archivePrefix={arXiv},
      primaryClass={cs.CL},
      url={https://arxiv.org/abs/2307.03172}, 
}
\bibliographystyle{acl_natbib}

\appendix

\section{Appendix}
\subsection{GPT-Annotated Utility Scores and Verification}
\label{sec:appendix-verification}
As part of our analysis, we also evaluate the upper bound performance achievable using GPT-4o annotated utility scores to rank passages. The results, summarized in Table~\ref{tab:gpt_upper_bound}, serve as a strong validation of our training data quality and support the effectiveness of reasoning-trace-based utility scoring.

As an additional check, we manually verified approximately 150–200 examples across the HotpotQA, MuSiQue, and 2WikiMultihopQA datasets. In over 95\% of cases, the LLM-generated utility scores aligned with human judgments of passage usefulness, confirming their reliability as supervision signals.

\begin{table*}[h]
\centering
\scriptsize
\begin{tabular}{lccccccc}
\toprule
\textbf{Dataset} & \textbf{P@2} & \textbf{R@2} & \textbf{F1@2} & \textbf{R@5} & \textbf{NDCG@1} & \textbf{NDCG@5} & \textbf{EM} \\
\midrule
HotpotQA & 93.29 & 93.29 & 93.29 & 98.39 & 96.99 & 95.85 & 89.77 \\
MuSiQue & 87.90 & 79.38 & 83.43 & 84.32 & 93.40 & 86.88 & 82.80 \\
2WikiMultihopQA & 98.20 & 86.32 & 91.87 & 98.85 & 99.80 & 98.92 & 78.96 \\
\bottomrule
\end{tabular}
\vspace{0.5em}
\caption{Upper bound performance using GPT-annotated utility scores for ranking.}
\label{tab:gpt_upper_bound}
\end{table*}

\subsection{Evaluating Decoder-only Models for Utility Scoring}
\label{sec:slm}
\begin{figure*}[t]
\centering
\includegraphics[width=0.9\textwidth]{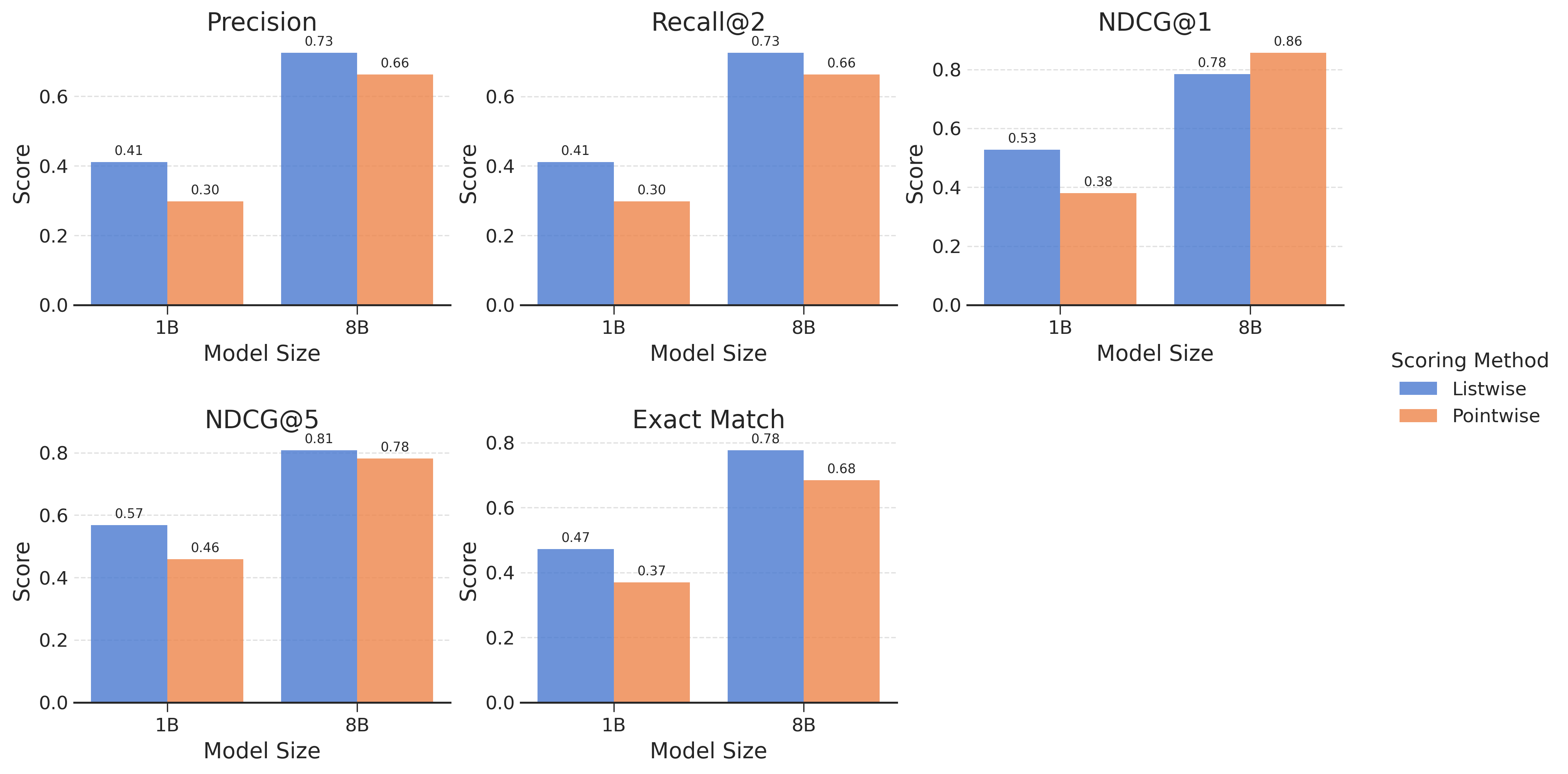}
\caption{\footnotesize Performance comparison of decoder-only models (LLaMA 3.2 1B and LLaMA 3.1 8B) on the HotpotQA dataset, fine-tuned using two different methods: Pointwise and Listwise scoring}
\label{fig:slm_comparison_1}
\end{figure*}
\begin{figure*}[t]
\centering
\includegraphics[width=0.9\textwidth]{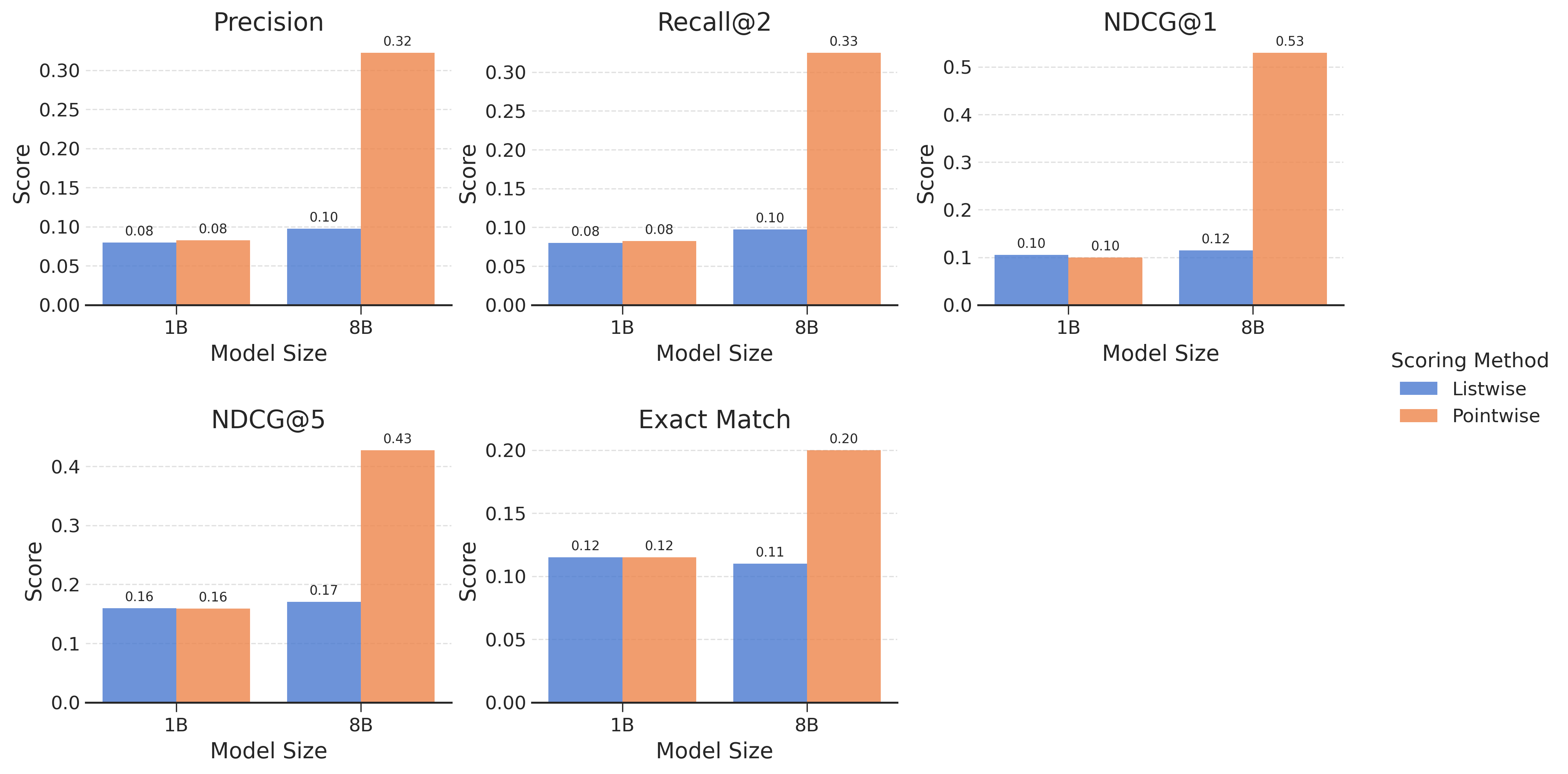}
\caption{\footnotesize Performance comparison of decoder-only models (LLaMA 3.2 1B and LLaMA 3.1 8B) on the MuSiQue dataset, fine-tuned using two different methods: Pointwise and Listwise scoring}
\label{fig:slm_comparison_2}
\end{figure*}
\begin{figure*}[t]
\centering
\includegraphics[width=0.9\textwidth]{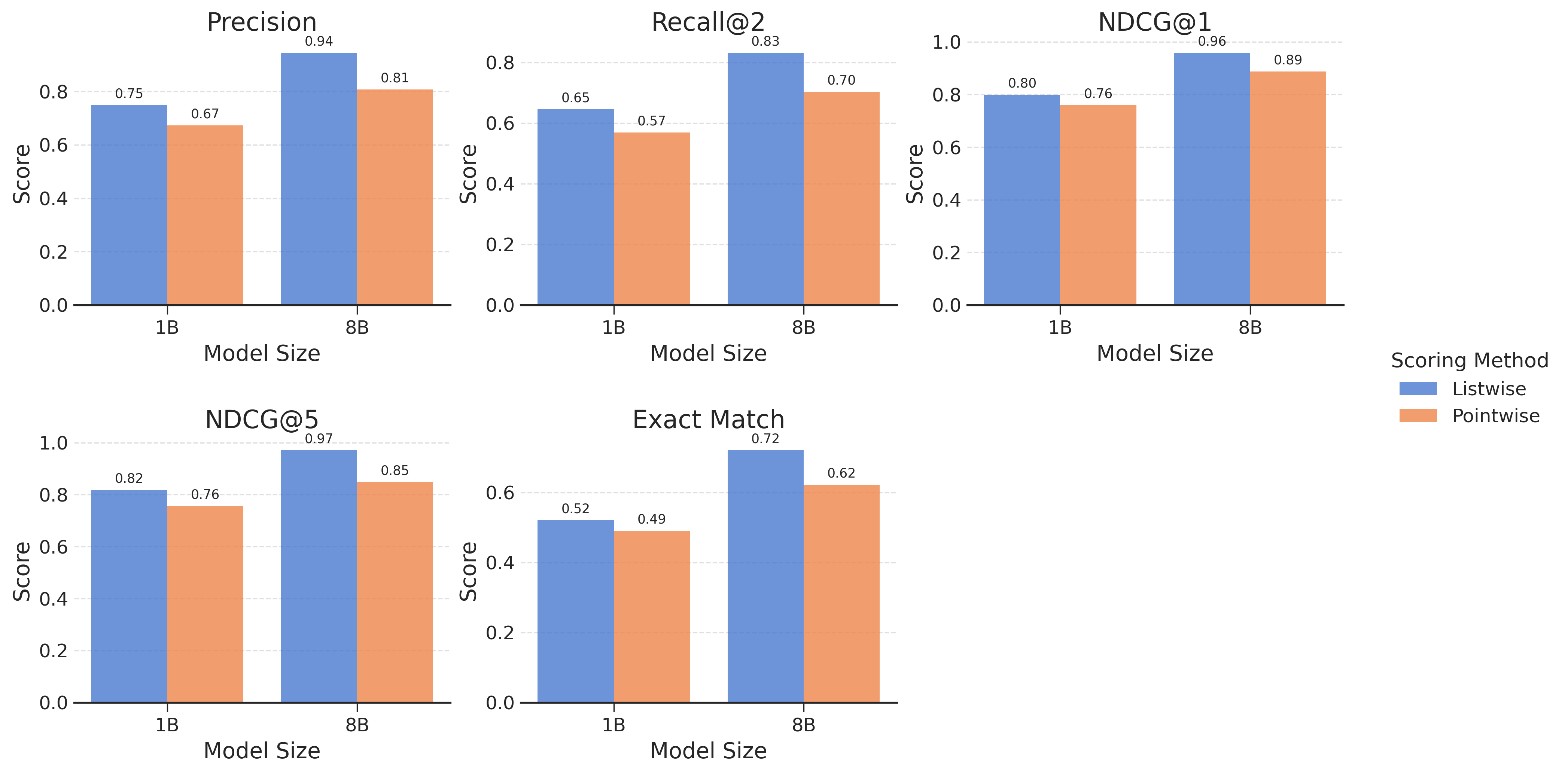}
\caption{\footnotesize Performance comparison of decoder-only models (LLaMA 3.2 1B and LLaMA 3.1 8B) on the 2WikiMultiHopQA dataset, fine-tuned using two different methods: Pointwise and Listwise scoring}
\label{fig:slm_comparison_3}
\end{figure*}

We conduct an auxiliary experiment by fine-tuning two decoder-only models, LLaMA 3.2 1B and LLaMA 3.1 8B \cite{meta2024llama3}, to compare their performance on this regression task against our primary encoder-only model. These models were fine-tuned using the same LLM-generated utility scores under two distinct settings:
(1) Pointwise Scoring: The model predicts a utility score for each passage individually, using a \texttt{(question, passage\k, utility\_score\_k)} triplet as input.
(2) Listwise Scoring: The model is presented with the question and the entire set of candidate passages simultaneously, aiming to predict individual scores with full context.
Figures~\ref{fig:slm_comparison_1} and  ~\ref{fig:slm_comparison_3} show that providing the decoder models with full context (listwise) offers only marginal benefits over the pointwise scoring for HotpotQA and 2WikiMultiHopQA. However, for the MuSiQue dataset (Fig ~\ref{fig:slm_comparison_2}), which has a larger candidate set (20 passages), the pointwise approach is significantly more effective. We hypothesize this performance degradation in the listwise setting is due to the well-documented Lost-in-the-middle \cite{baker2024lostmiddleinbetweenenhancing, liu2023lostmiddlelanguagemodels} problem, where decoder models struggle to attend to all items in a long input sequence.
It is also noteworthy that the absolute performance of these fine-tuned LLaMA models is mostly lower than our lightweight RoBERTa-based model (as reported in Table~\ref{tab:contextual-utility-eval}). Despite having a more constrained input during inference (scoring each passage individually), our model is better at identifying the useful passages and correctly ordering them, further underscoring the quality of our training data.

\end{document}